\documentclass[final]{cvpr}

\usepackage{times}
\usepackage{epsfig}
\usepackage{graphicx}
\usepackage{amsmath}
\usepackage{amssymb}
\usepackage{booktabs}
\usepackage{makecell}
\usepackage{adjustbox}
\usepackage{balance}

\usepackage[pagebackref=true,breaklinks=true,colorlinks,bookmarks=false]{hyperref}

\def\cvprPaperID{7} 
\def\confYear{CVPR 2021}

\begin{document}

\title{Adaptive Intermediate Representations for Video Understanding}


\author{
Juhana Kangaspunta, AJ Piergiovanni, Rico Jonschkowski, Michael Ryoo, Anelia Angelova\\
Google Research\\
{\tt\small \{juhana,ajpiergi,rjon,mryoo,anelia\}@google.com}
}

\maketitle

\begin{abstract}

   A common strategy to video understanding is to incorporate spatial and motion information by fusing features derived from RGB frames and optical flow. In this work, we introduce a new way to leverage semantic segmentation as an intermediate representation for video understanding and use it in a way that requires no additional labeling.                                                       
    Second, we propose a general framework which learns the intermediate representations (optical flow and semantic segmentation) jointly with the final video understanding task and allows the adaptation of the representations to the end goal. Despite the use of intermediate representations within the network, during inference, no additional data beyond RGB sequences is needed, enabling efficient recognition with a single network.           
    Finally, we present a way to find the optimal learning configuration by searching the best loss weighting via evolution.                                   
    We obtain more powerful visual representations for videos which lead to performance gains over the state-of-the-art.         

\end{abstract}

\section{Introduction}



In the task of action recognition, the model is given only a video-level action class as a supervision signal. From this, the model has to implicitly learn a wide variety of concepts from the input data: objects, people, motion and their relationships through time. Thus using explicit intermediate representations can aid in the learning process. One notable form of intermediate representation, optical flow, 
has been used extensively in two-stream architectures~\cite{simonyan2014two,carreira2017quo,feichtenhofer2016convolutional}. Optical flow is often pre-computed using an algorithm optimized for the quality of optical flow itself, rather than the downstream video understanding task.

\begin{figure}[t]
  \centering
  \includegraphics[width=0.5\textwidth]{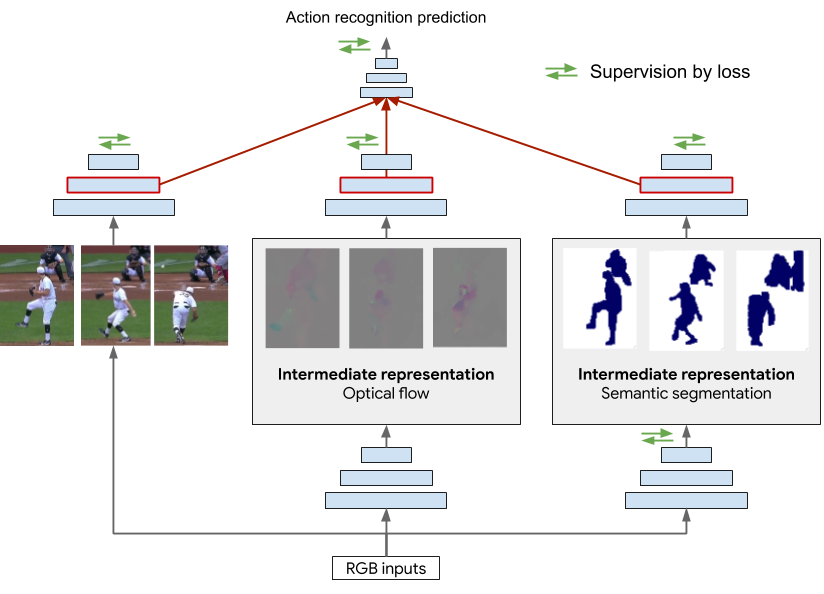}%
  \caption{
  Intermediate representations from optical flow (middle) and semantic segmentation (right) are learned alongside the main RGB task and their losses are simultaneously optimized for improving the performance of the main video understanding task. Merging of streams is highlighted with red borders and supervision by loss is highlighted in green.}
  \label{fig:teaser}
\end{figure}

As an addition to motion information, knowledge about the presence and location of people and objects should also be highly relevant in understanding the activity in a video sequence~\cite{jain2015what}. We propose a new way to use semantic segmentation as an intermediate representation learned by the network (Figure \ref{fig:teaser}). Semantic segmentation is learned along with the main video understanding task using annotations generated by an off-the-shelf teacher model that was trained on static images unrelated to the task. Therefore, the annotations are readily available and require no additional labeling effort.

How can one best take advantage of these intermediate representations?
Enforcing \emph{rigid} intermediate representations, for example by training and freezing a model for optical flow, prevents these representations from learning information that is specific to the given task, e.g., when some types of motion are more informative than others. To allow for such flexibility, we encourage intermediate representations in a ``soft'' way through auxiliary learning objectives that are optimized jointly with the video understanding task. Importantly, after training the network and learning the corresponding intermediate representations, the network requires RGB-only inputs at inference. 

As all intermediate representations are jointly learned with the final task by optimizing a weighted sum of these objectives, the weights of their losses determine the importance of each stream's feature representations and intermediate losses. In addition to optimizing these representations, we also optimize the loss weights for end-to-end action recognition performance using evolution to maximize mutual usefulness of each loss (Figure~\ref{fig:teaser}). 
We call the approach `AIRStreams', short from Adaptive Intermediate Representation Streams, as it provides a  framework for using different streams of intermediate representations in an adaptive manner for video understanding.

Our contributions are threefold: (a) We propose to incorporate semantic segmentation as a learned intermediate representation for video understanding that is complementary to optical flow, improves performance, and does not require additional labeling.
(b) We provide a novel framework that combines different intermediate representations in a way that automatically optimizes them with respect to the final task. The networks in our framework require RGB inputs only. c) We use an evolutionary strategy to further balance the learning of these representations for final action recognition task.

Our experimental evaluation shows that AIRStreams performs well across three major action recognition datasets. In an extensive analysis of the effect of adaptive intermediate representations and the proposed approach, we find that the extra modalities help, more specifically: (i)  Adding the semantic segmentation stream significantly improves performance, and can do so without pre-training. 
(ii) Intermediate representations are more useful when optimized jointly with the task by allowing the end-to-end gradient to affect the intermediate representation. (iii) Optimizing the weights of the intermediate losses for the task provides additional improvements in performance. (iv) Combining the aforementioned concepts, AIRStreams outperforms or is competitive with the best current video understanding models in multiple action recognition datasets.

While incorporating extra modalities requires more computation, we opted for an
orthogonal approach where we use a lightweight backbone, 
but utilize the compute to pack additional modalities, which
bring in new valuable information and large improvements. At inference, our
model is very efficient, taking 17 GFLOPs, as opposed to SlowFast~\cite{feichtenhofer2018slowfast} which is 213 GFLOPs, and I3D~\cite{carreira2017quo} - 216 GFLOPs.

\begin{figure*}[t]
  \centering
  \includegraphics[width=0.82\textwidth]{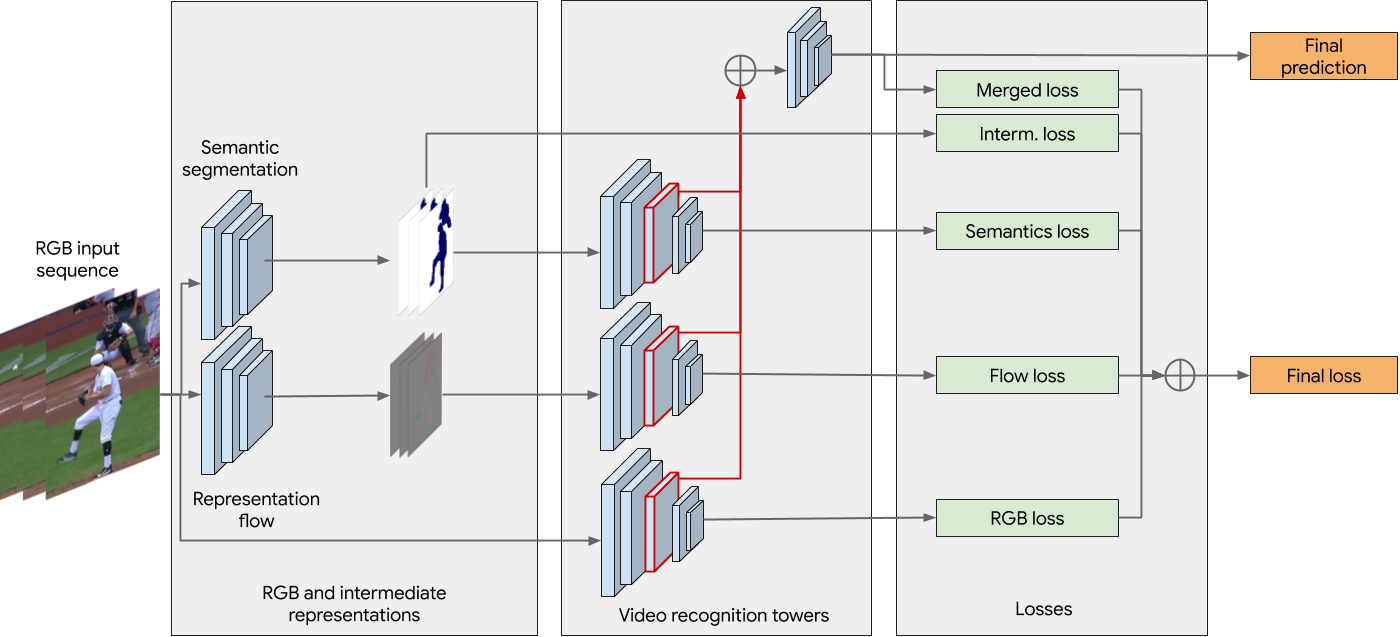}%
  \caption{
  AIRStreams overview: The input RGB video gives rise to intermediate representations based on semantics (top stream), where common objects are automatically segmented, and optical flow (middle stream). Together with the RGB stream (bottom stream), each of the representations are passed into their own video recognition towers. Features extracted from the middle of each video recognition tower are merged (highlighted with red borders) and passed through an additional convolutional tower to produce the final action recognition prediction. The losses (highlighted in green) of each stream and the intermediate losses are simultaneously optimized for improving the performance of the main video understanding task.
  }
  \label{fig:overview}
\end{figure*}

\section{Previous work}

Understanding videos has had many successful approaches applying a number of deep architectures~\cite{wang2011action,tran2014c3d,simonyan2014two,Feichtenhofer2016Spatiotemporal,Wang2016apperance,kay2017kinetics,xie2018rethinking,carreira2017quo,tran2018closer,fan2018e2e,feichtenhofer2018slowfast}. 
Optical flow, although a product of the raw video stream, has been demonstrated to be of high value to video recognition~\cite{sevillalara2018on}.
Two-stream networks~\cite{simonyan2014two,carreira2017quo,feichtenhofer2016convolutional} are particularly interesting constructs with RGB and flow streams, in which optical flow is used as a separate  input to a deep network, instead of relying on the RGB sequence to learn flow information on its own.  
Early versions, e.g., Wang et al.~\cite{wang2016temporal}, used an RGB difference stream and a warped RGB stream after applying flow.
While the majority of two-stream networks apply late-fusion, by fusing features from the two streams before the final output, the work of ~\cite{Feichtenhofer2016Spatiotemporal} proposes fusion across layers from multiple streams.

Slow-fast networks~\cite{feichtenhofer2018slowfast} consider two individual RGB-only streams, 
which jointly process information at different temporal resolutions. They also advocate for fixed connections between parallel layers.
We here propose the semantics stream and to optimize the performance and mutual benefit of multiple streams working jointly.
The above-mentioned works can benefit from AIRStreams by incorporating extra semantic streams, by training end-to-end, and learning the importance of the loss contributions of each of the streams.

Some previous approaches have successfully used object information for video understanding, e.g.~\cite{xu2015can,kalogeiton2017joint,ji2018e2e,ray2018SOA,jain2015what,wang2016two,mettes2017spatial,baradel2018object,videoasgraph}.
Sigurdsson et al.~\cite{sigurdsson2017what} demonstrates the potential of using objects via object labels oracles. 
Other works have explored mutli-steam networks with various inputs. For example, PoTion \cite{choutas2018potion} proposes a complementary pose representation. Chained Multi-stream Networks  \cite{chained} similar to this work uses multiple different input features each optimized for recognition. This paper instead uses other features, like objects, to learn representations within the network so that during inference they are not needed.
While we are not aware of prior work using semantics as a jointly optimized separate stream, Diba et al.~\cite{diba2019holistic} is most related to ours. They utilize a 2D stream which is built from image pretraining on an independent dataset. It is infused into a 3D stream which processes the main video information. 


Multi-task learning approaches~\cite{doersch2017multi,ren2018cross,zamir2018taskonomy,kendall2018multi,hsu2018unsupervised,munro2020multimodal} train several tasks in parallel (each one to a corresponding loss) and have been successfully shown to use  multiple tasks for mutual benefit, e.g., for joint feature embedding learning. 
In another aspect of multi-task learning, various modalities (or streams) are considered as tasks but the focus is on training towards separate output goals, usually called `heads', while maintaining shared representations, e.g., learning by supervising for depth, surface normals etc., while learning image segmentation.
Other works also do joint optimization in the video understanding context, e.g. 
SegFlow ~\cite{cheng2017segflow} of optical flow and video object segmentation.
Unlike the multi-task learning approaches listed above, we train the intermediate representations to help the main action recognition task and have the main task be explicitly dependent of the intermediate representations, not only indirectly dependent through feature sharing. Previous works have found that optimizing flow in conjunction with recognition improves performance \cite{ng2018actionflownet,fan2018end}.


Architecture search~\cite{zoph2017neural,zoph2018nas,liu2018progressive,bender2018understanding,liu2019darts,real2019amoeba,hu2019efficient} 
has proven to be highly successful for visual understanding. 
Recent work on architecture search for videos~\cite{nekrasov2019architecture,piergiovanni2018evolving} has demonstrated promise. Evolutionary search is also used for selecting a subset of losses for unsupervised feature learning~\cite{piergiovanni2020evolving}.
Our work too uses evolutionary search for loss combinations, where evolution is used as a versatile and easy-to-apply optimization technique, especially for non-differentiable tasks such as ours. While other optimization techniques could substitute evolution, 
we are not aware of prior work which proposed to optimize intermediate representations for video understanding and jointly optimize all representations contributions to the final task.






\section{Learning with adaptive intermediate representations}

In this section, we describe our video understanding model, the different modules it consists of, and the loss functions for training. 

\subsection{Model overview}

The AIRStreams model consists of three independent streams of computation (Figure~\ref{fig:overview}). An RGB stream is optimized only towards the final action recognition task without any intermediate constraints. The two other streams, in addition to being optimized towards the final task, each produce an intermediate output: optical flow and semantic segmentation. Here we first explain the overall model architecture, the individual streams in detail, how the features from the streams are merged together, what losses are used during training and how these losses are weighted.

\paragraph{Model architecture.}
The RGB video input and each individual intermediate representation are processed by a separate convolutional action recognition tower, e.g., (2+1)D ResNet~\cite{tran2018closer}. These towers are shown in Figure~\ref{fig:overview} in the middle section. Different tower backbones can be used and we experiment with a lightweight alternative in this paper. Each of the individual towers attempt to predict the final task and are supervised by separate classification losses. The feature maps of each of the towers are also fused together to produce a merged stream, which has a classification loss as well. The merged stream produces the final output that is used in evaluation. More advanced connectivity between the towers is also possible, e.g., as in~\cite{Feichtenhofer2016Spatiotemporal} or in RandWire Networks~\cite{xie2019randwire}. 

Since it is not immediately clear which configuration will work well, we define below a set of losses (Section~\ref{sec:losses}) and propose to further optimize them automatically via evolution (Section~\ref{sec:evo}). 

The way in which each intermediate representation is produced is specific to the representation. The framework is extendable to a variety of different representations, which can be unsupervised or supervised. Below, we describe the computation of the intermediate representations in detail.

\paragraph{RGB Stream.}
The RGB stream takes as input a sequence of RGB images from a video and applies a convolutional tower to it. This is the standard approach for many video understanding tasks.

\paragraph{Segmentation Stream.}
One of the key aspects of AIRStreams is using semantic segmentation as an intermediate representation. Our rationale is that the presence and location of people and semantic objects hold key information for the activities at hand. 
Importantly, the interaction of the features with others (flow or RGB) as a result of the presence of the semantic stream in the AIRStreams architecture will provide opportunity for better learning of the final actions.

To produce the semantic segmentation representation, we use a Dilated ResNet-18 \cite{yu2017dilated} with output stride 16 applied frame-by-frame on the input RGB sequence. We use pixelwise softmax cross-entropy as the semantic segmentation loss using model-generated annotations as ground truth as explained below. The pre-softmax activations of the semantic segmentation network are then used as the input to a subsequent representation-specific action recognition tower. 

Frame-by-frame annotation of object segments is very expensive and therefore large-scale action recognition datasets do not provide such annotations. The semantic annotations used to supervise the network explained above are obtained with an off-the-shelf algorithm (trained on an independent dataset of static images, e.g., on MS-COCO dataset~\cite{mscoco}), which are readily available, and incurs no further labeling effort or cost. Notably, this teacher model used to generate the semantic segmentation annotations is a much larger and much more complicated Mask R-CNN model \cite{he2017rcnn} compared to the network generating semantic segmentation inside AIRStreams.

Many of the classes present in MS-COCO are likely not encountered in action recognition datasets, e.g., giraffe, kite, etc., but a handful of classes will be useful and prevalent in the videos such as people, furniture and objects of daily life. We conjecture that the semantic features will encode valuable information, as well as directly transfer information for some of the classes that are jointly encountered - e.g., person, chair, etc.

\paragraph{Optical Flow Stream.}
While many video understanding models use a single-stream, two-stream models~\cite{simonyan2014two,carreira2017quo} (RGB and optical flow), have become increasingly popular. The rationale is that flow directly provides information about motion in a much more efficient way than extracting it from the raw RGB sequence. This approach has been shown to be very beneficial. Our approach includes an optical-flow based intermediate representation as well.

Our use of optical flow has two key aspects.
First, in order to learn and fine-tune the flow representation for the final task, we need a fully differentiable version of flow. We show later in experiments that backpropagating from the final classification through the intermediate representation computations is important. For a fully differentiable flow we use Representation Flow~\cite{piergiovanni2018representation}, which is trained on-the-fly and can be backpropagated fully. 
Representation flow mimics the implementation of the variational flow algorithms~\cite{brox2004high} via fully differentiable learnable layers, minimizing the total variational energy. We chose this over approaches like FlowNet/ActionFlowNet \cite{ilg2017flownet,ng2018actionflownet} as this requires no pre-training to obtain good flow representations.

Second, similar to the semantic segmentation representation, the optical flow is passed into its own action recognition tower. Unlike regular two-stream towers, which all apply late fusion, features from this video recognition tower are fused early with the other video recognition towers as shown in Figure~\ref{fig:overview} and supplementary materials, as well as given its own video classification loss. 

While we use representation flow for our experiments, other versions of flow can be incorporated. 

\paragraph{Merging streams.} \label{sssection:merging}
The features from each of the three action recognition towers (flow, semantic segmentation and RGB) are combined via averaging at a pre-defined level. We merge after the $3^{rd}$ block and our net has a total of $6$ blocks. After the features have been merged, the rest of the video recognition tower for the merged stream is identical in structure for blocks $4$ to $6$ of the individual feature towers. Similarly as for the feature action recognition towers, the merged tower is supervised by the final action recognition task.
Merging at more than one levels (including the common late-fusion) is also possible and can easily be added to the AIRStreams framework as separate losses.

\subsection{Losses} \label{ssection:losses}
\label{sec:losses}

In AIRStreams, we propose to use multiple independently-weighted losses, described below. 

\paragraph{Activity recognition losses.}
Each action recognition tower, one consuming flow, one consuming semantic segmentation, one consuming RGB and one consuming the merged features, has an individual softmax cross-entropy (SCE) loss $L_{rgb}$, $L_{flow}$, $L_{semantic}$, $L_{merged}$ with respect to the final action recognition target:
$L_{*}=SCE(Y_{*}, Y_{target})$, where $Y_{target}$ are the target labels for the final activity recognition task, and the $*$ denotes each of the output streams' logits, i.e. $_{rgb}$, $_{flow}$, $_{semantic}$, $_{merged}$.

\paragraph{Intermediate representation losses.}
Additionally, each intermediate representation can have its own loss that is optimized for the specific task. In this work, we have semantic segmentation and optical flow as intermediate representations, from which only semantic segmentation requires its own loss. We use standard pixelwise softmax cross-entropy (SCE) between the logits $Y_{interm\_semantic}$ and the `target' semantic labels $Y_{labels\_semantic}$ as the loss for semantic segmentation $L_{interm}$. It is applied per frame. Note that the `ground truth' labels used for this loss are generated by an off-the-shelf segmentation model trained on the MS-COCO dataset~\cite{mscoco} with 90 classes. Therefore, we do not require pixelwise labeling of the action recognition datasets.
\begin{equation}
    L_{interm} = SCE(Y_{interm\_semantic}, Y_{labels\_semantic}).
\end{equation}

\paragraph{Final loss.}
Putting it all together, we have a number of losses from each of the action recognition streams (flow, semantic segmentation, RGB and merged) and an intermediate representation loss for semantic segmentation. The losses are naturally weighted by parameters $\lambda_{rgb}, \lambda_{flow}, \lambda_{semantic}, \lambda_{merged}, \lambda_{interm}$ which determine the relative importance of each sub-task.
\begin{equation}
\begin{split}
L = \lambda_{rgb}L_{rgb} + \lambda_{flow}L_{flow} + \lambda_{semantic}L_{semantic} + \\ \lambda_{merged}L_{merged} + \lambda_{interm}L_{interm}
\end{split}
\end{equation}

\subsection{Evolution of loss weighting}
\label{sec:evo}

AIRStreams can be viewed as a collection of losses which train jointly for the final action recognition task. With that, its de-facto topology can depend on which losses are switched on or off (via their weights). 
For example, presence or absence of certain losses can make a big difference, as Section~\ref{sec:ablations} shows accuracy of only $42.82$ is achieved on HMDB for uniform weights (all weights are 1s) vs. $50.02$ when the flow loss and the semantic losses are both 0s (all others 1s).

We propose to learn the combination of losses via {\it evolution} of weights for the purposes of achieving a higher performance on the main task \cite{piergiovanni2020evolving}.
Evolution~\cite{goldberg91acomparative,real2019amoeba} is a well established technique in which a random set of individuals in a population (here combination of weights) are generated, after which random mutations are applied. Their performance is tested and the best performing combination of weights are preserved for the next iterations.

For AIRStreams, we use a tournament selection evolutionary strategy \cite{goldberg91acomparative}. For this experiment, we use a population size 10 and tournament size 3. At each evolutionary step, we mutate an individual weighting by randomly selecting one weight and assigning it a random value between 0 and 1. The evolution evaluates 100 different loss weightings in total.

This approach can be viewed as an automatic optimization of weights, in lieu of parameter tuning. 
We note that the number of losses is likely to increase with more advanced tasks and representations, thus tuning them by hand or exhaustive search will be a prohibitively expensive task in such scenarios.

\subsection{Inference}
 We measure end-to-end performance for the final action recognition task by using the output of the merged stream. We note that RGB is the only input and no pre-computed flow or semantics information is needed. Despite computing multiple towers corresponding to `modalities', our model is very fast with 17 GFLOPs per frame, much fewer than SlowFast~\cite{feichtenhofer2018slowfast} with 213 GFLOPs, and I3D~\cite{carreira2017quo} with 216 GFLOPs.

\subsection{Network backbone}
Instead of using a standard ResNet-50 or ResNet-101 architectures, we here experimented with a more lightweight model, so that we can take advantage of multiple towers. 
The network architecture backbone we are using is a Tiny Video Network~\cite{piergiovanni2019tvn}, and open source code~\cite{piergiovanni2020tvntfhub,piergiovanni2020tvncode}, which is comparable to
a ResNet-18.
The network architecture itself contains 6 blocks with 1d convolution and spatial convolutional layers, and efficient blocks, such as context gating~\cite{xie2018rethinking} and squeeze-and-excite~\cite{hu2018squeeze}; it is described in more detail in the supplementary materials. This model is of lesser capacity to standard architectures for video understanding e.g., ResNet-50, but it is more efficient to train so we prefer it for faster convergence. Importantly, it allows us to perform well without pre-training on small datasets. Of note is that it is less powerful than a ResNet-50, as evidenced by the performance without pre-training (see baseline models in Table~\ref{tab:evo}). Despite that, within the AIRStreams framework, it outperforms the best state-of-the-art approaches on very challenging datasets (as seen later, e.g., in Table~\ref{tab:mit}).

\section{Experiments and results}
\label{sec:ablations}

First we introduce the datasets we used. In the subsequent sections, we investigate multiple ways intermediate representations impact video classification tasks and how different components contribute to the performance of AIRStreams.

\subsection{Datasets}

\noindent \textbf{Moments in Time (MiT)}. The MiT dataset~\cite{monfort2018moments} is a large scale dataset which consists of about 1M videos, each one approximately 3 seconds long. It has 339 classes,
802,264 training, 33,900 test videos.

\noindent \textbf{Charades dataset}. The Charades~\cite{sigurdsson2018charadesego} dataset is a multi-class multi-label dataset of 157 activities. It has 7,985 training and 1,863 test videos. We follow the standard protocols for evaluation for multi-class multi-label setting and thus use a multi-class sigmoid cross-entropy classification loss instead of a softmax cross-entropy loss for this dataset. 

\noindent \textbf{Toyota Smarthomes dataset} The Toyota Smarthomes dataset~\cite{das2019toyota} consists of daily lives activities, e.g., making coffee, reading, watching TV. It contains 16,115 videos and 31 action classes. 

\noindent \textbf{HMDB Video dataset}. HMDB~\cite{kuehne2011hmdb} is a smaller video recognition dataset with about 7,000 videos and 51 activities.
We use this dataset for our ablation experiments.


\subsection{Analysis of AIRStreams}
We start by investigating various aspects of learning adaptive intermediate representations for AIRStreams.

\vspace{-0.3cm}
\subsubsection{Intermediate representations help the final task}

We first examine how additional intermediate representations help in the action recognition task. We evaluate the use of flow, semantics and both, in addition to an RGB stream.

Table~\ref{tab:interm} shows the results. We can see that each of the intermediate representations help, and that the use of each of the streams alone is highly beneficial. When combined, they increase the overall performance even further for an overall improvement of above 8.3\% in absolute value, which constitutes about 20\% relative improvement.
These results are obtained when respective intermediate representation are optimized for the downstream task at hand. They are also conducted without pre-training from another dataset.
Interestingly, we can see that the effects of the semantic stream are very positive, despite the fact that the semantic labels are trained on an independent dataset from an off-the-shelf algorithm, and not related to the data on hand.

\vspace{-0.3cm}
\subsubsection{Intermediate representations should be optimized for the final task}
One of the key aspects of the AIRStreams architecture is learning adaptive intermediate representations jointly with the task at hand. This is in contrast to many prior works, which use pre-computed versions of flow. 
In this section we investigate whether optimizing intermediate representations for the final task is of importance. 
In order to evaluate this, we test two configurations. In both configurations, we train the intermediate representations jointly with the main task. However, in one configuration, we don't let gradients propagate from the final action recognition classification losses through the intermediate representation computations. This means that the intermediate representations are independent from final task and, for example, the semantic segmentation network is only optimizing towards the semantic segmentation pixelwise cross-entropy loss. In the other configuration, we let gradients propagate freely. This means that the intermediate representations are allowed to adjust for the final task, balancing the rigidity of the intermediate form with the needs of the downstream task.

Table~\ref{tab:stop_hmdb} shows the corresponding ablation study. We see that full optimization of the intermediate representations (or combination of them) for the task at hand is better than using pre-computed intermediate representations.
Each task is able to gain from aligning its intermediate features to the final task at hand.
While this result is not surprising, it 
solidifies the insight that one needs to optimize jointly, rather than pre-compute.
Importantly, backpropagating with the semantics labels brings more improvements, even though the semantic information is transferred from another task. 
Figure~\ref{fig:interm} shows why -- some objects which were missed by the segmentation model, can be recovered during training.  



\begin{table}
\centering
\adjustbox{width=0.95\linewidth}{
\begin{tabular}{lcc}
\toprule
Method & Acc. (\%)  & Improvement \\  
\midrule
RGB          & 41.72  & - \\
RGB + Semantics    & 46.32  & +4.6      \\
RGB + Flow          &47.66 & +5.94 \\
RGB + Semantics + Flow    & \textbf{50.02}   & \textbf{+8.3} \\
\bottomrule
\end{tabular}}
\caption{Ablation study. 
Intermediate representations improve performance, for both semantics and flow intermediate representations. 
Combining them helps most, with an 8.3\% improvement. 
HMDB dataset, training is done from raw data (no pretraining), to isolate other effects.  
The middle column shows classification accuracy on the dataset. The right column shows the corresponding improvements from RGB only.
}
\label{tab:interm}
\end{table}

\begin{table}
\begin{center}
\adjustbox{width=0.95\linewidth}{
\begin{tabular}{lcc}
\toprule
Learning mode & \makecell{Stop gradient\\at interm. repr.} & \makecell{Propagate\\gradient} \\
\midrule
RGB (no interm. repr.)                     &41.72            &-       \\
\midrule
RGB + Semantics           &43.57            &\textbf{46.32}   \\
RGB + Flow                &46.84            &\textbf{47.66}   \\
RGB + Semantics + Flow      &47.76            &\textbf{50.02}   \\
\bottomrule
\end{tabular}}
\caption{Ablation study for intermediate representations for HMDB
All intermediate representations are beneficial (compare column-wise to RGB only).
For every intermediate representation, allowing it to be learned end-to-end is more beneficial than using a pre-specified representation (compare row-wise). Accuracy in \% shown for both columns.
}

\label{tab:stop_hmdb}
\end{center}
\end{table}

\begin{table} 
\begin{center}

\adjustbox{width=0.95\linewidth}{
\begin{tabular}{lcccc} 
\toprule
Dataset & \makecell{No aux.\\losses} & \makecell{Fixed\\weights} & \makecell{Evolved\\weights} \\
\midrule
HMDB (no pretr.)     &41.72    & 50.02      &  \textbf{50.53}       \\
Charades (no pretr.)       & 28.56       &35.05        & \textbf{35.45}    \\

\bottomrule
\end{tabular}}
\caption{Ablation study: Performance when evolving weights for balancing all losses from all intermediate representations, compared to manually tuned loss combinations.
Accuracy(\%) for HMDB and mAP(\%) for Charades are shown.}
\label{tab:evo}
\end{center}
\end{table}

\vspace{-0.4cm}
\subsubsection{Learning how to combine intermediate feature representations.}
From the above-mentioned ablation results, we learned that (i) intermediate representation are beneficial and their combination is of highest value and (ii) they should be optimized jointly with the final task. The above experiments are run with fixed weights of 1 on `merged', `interm' and `RGB' and 0s for flow and semantic tower, which is already a very successful model (close to the evolved coefficients). 
In this subsection, we run ablation experiments to evaluate the evolution of the weights of auxiliary losses via the evolution procedure. More specifically we compare: the baseline model without intermediate representations, the model with all intermediate representations with fixed loss weights, and the full AIRStreams evolved model, where the weights are additionally evolved for the task at hand. 

Table~\ref{tab:evo} compares the performance from weights obtained by evolution vs fixed weights vs a baseline with no intermediate weights. Here the gradients are allowed to flow so that full learning and optimization is done.
We observe that the evolved weights give further boost, obtaining the best performance. The improvement by evolution is small but consistent, as also observed across our experiments. This is due to the initial good choice of fixed coefficients; other trivial combinations are not as good, for example using \textbf{uniform weights}, achieves $42.82$ on HMDB, which is above the baseline, but much worse than $50.02$ and $50.53$, with evolution.
We note that the evolved weights applied to the Charades dataset are evolved on a different dataset (MiT). 
In summary, for both datasets, AIRStreams improves performance over the baseline by large margins. 
The best evolved loss weights are displayed in Figure~\ref{fig:evo_weights}.

\subsubsection{Model visualizations}

\begin{figure}
  \centering
    \includegraphics[width=1.01\columnwidth]{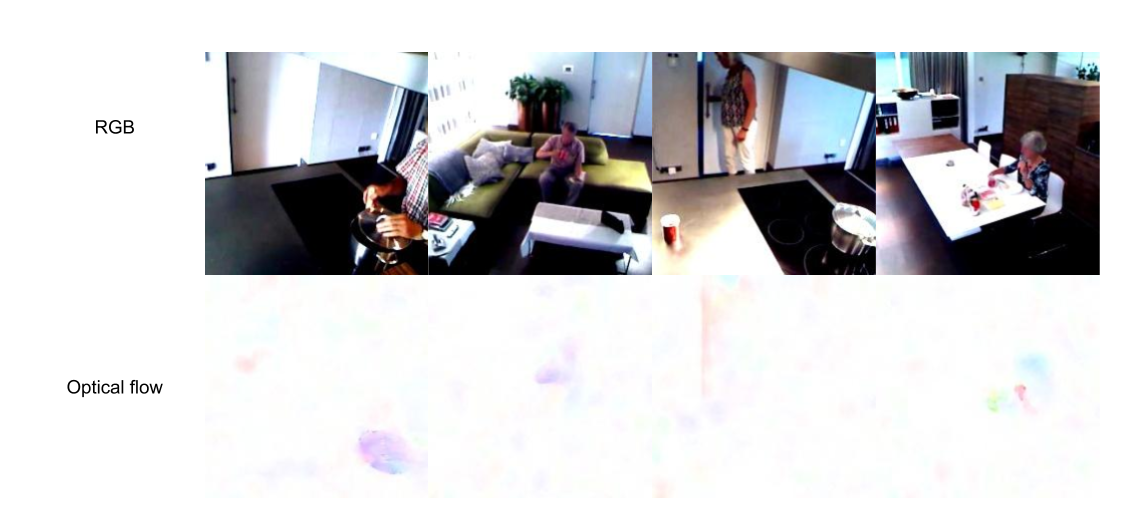}
  \caption{Samples from the learned intermediate representations for optical flow. The optical flow is learned as a part of the overall training process and gradients are backprogated from video classification losses allowing the flow to be tuned for the end task.}
  \label{fig:interm-flow}
\end{figure}

\begin{figure}
  \centering
    \includegraphics[width=1.01\columnwidth]{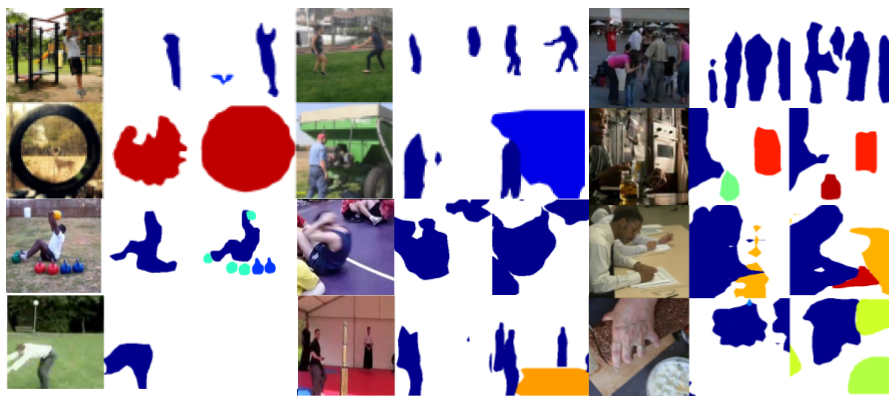}
  \caption{Samples from the learned intermediate representations for semantics. 
  These are the class labels perceived at the input point of the semantic tower.
  Each panel contains image, learned segmentation, and `ground truth', the latter being trained on another dataset and may miss objects or be incorrect. The learned segments are less sharp, but end-to-end training is able to correct errors in the `ground truth' by recovering full segments, e.g., bottom left. 
  Interestingly, the majority of segments are persons (dark blue), other classes may appear too. Best viewed in color.}
  \label{fig:interm}
\end{figure}

Figures~\ref{fig:interm-flow} and~\ref{fig:interm} visualize the learned intermediate representations.
Figure~\ref{fig:interm} shows intermediate representations for semantics. The visualization shows the output of the teacher model as well as the intermediate representations learned as a part of the AIRStreams training.
In order to show the visualization, we use an argmax over the logits, whereas during training, actual logits are passed to the video recognition network. We observe that the learned representations are often less sharp, but in some cases the training is able to correct errors in the segmentation algorithm (mostly by proposing a missing segment, as in the bottom left example). We conjecture that letting gradients flow through from the final action recognition task, the network is able to learn additional elements not present in the frame-by-frame data from the teacher model.



\begin{figure}
  \centering
  \includegraphics[width=0.49\columnwidth]{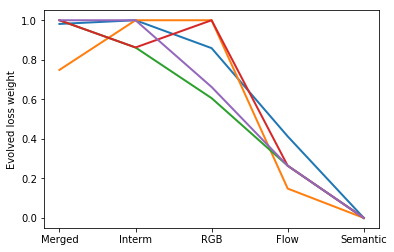}
  \includegraphics[width=0.49\columnwidth]{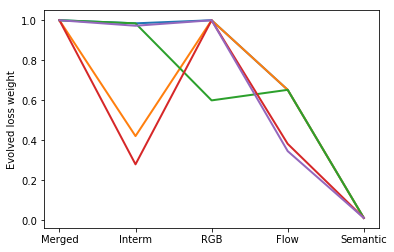}
  \caption{Learned best coefficients after evolution. Evolved on MiT (left), HMDB (right).}
\label{fig:evo_weights}
\end{figure}

Figure~\ref{fig:evo_weights} visualizes the loss coefficients learned from the evolutionary search. The figure shows the learned top-5 coefficients on HMDB and MiT. As seen, 
the evolution tends to assign larger weights to merged and the main RGB towers, and to the semantics intermediate representation loss, whereas smaller weights to the flow and semantics action recognition towers.
While multiple evolution runs will produce different sets of coefficients, the final performance 
is similar for top performing evolved coefficients. For example, accuracy on MiT with top coefficients evolved on MiT is $34.51$ vs $34.44$ when evolved on HMDB. 


\subsection{Comparison to previous methods}

Here we compare AIRStreams to the state-of-the-art on various action recognition datasets.




\begin{table} [t]
\begin{center}
\adjustbox{width=0.99\linewidth}{
\begin{tabular}{llc}
\toprule
Method & Mode   & \makecell[c]{Acc.(\%)} \\                         
\midrule
TSN~\cite{wang2016temporal}  & RGB & 24.11  \\
\makecell[l]{ResNet50-ImageNet~\cite{carreira2017quo} (RN50)} & RGB & 27.16  \\
TRN-Multiscale~\cite{zhou2018temporal} & RGB & 27.20  \\
\makecell[l]{Spatio-temporal attention\cite{meng2019interpretable}} &RGB    &27.86 \\
\midrule
TSN-2Stream~\cite{wang2016temporal} & RGB+Flow & 25.32  \\
TRN-Multiscale~\cite{zhou2018temporal} & RGB+Flow & 28.27  \\
I3D~\cite{carreira2017quo} & RGB+Flow & 29.51  \\
EvaNet~\cite{piergiovanni2018evolving}  &RGB+Flow  &30.5 \\
\makecell[l]{EvaNet~\cite{piergiovanni2018evolving} (Ensemble)}  &RGB+Flow  &31.8 \\
AssembleNet~\cite{ryoo2020assemblenet} RN50 &RGB+Flow  &31.41 \\
AssembleNet~\cite{ryoo2020assemblenet} RN50 &RGB+Flow (+Kin.) &33.91 \\
AssembleNet~\cite{ryoo2020assemblenet} RN101 &RGB+Flow (+Kin.) &34.27 \\
\midrule

\makecell[l]{AIRStreams  (Ours)}  &RGB+Flow     & 33.53    \\
AIRStreams (Ours)  & \makecell[l]{RGB+Flow+Sem.}   & \textbf{34.51}      \\
\bottomrule
\end{tabular}
}
\caption{Performance on the Moments-in-Time (MiT) dataset. 
AIRStreams is able to outperform the SOTA (state-of-the-art) on this challenging dataset, both with flow-only and with semantics+flow, and \textbf{without using any pre-training}.
}
\label{tab:mit}
\end{center}
\vspace{-6mm}
\end{table}


\vspace{-0.3cm}
\subsubsection{Moments in Time dataset} We evaluate on the challenging large-scale Moments-in-time dataset~\cite{monfort2018moments}, comparing to prior methods including 2-stream methods (RGB and flow), most of which use pre-training (Table~\ref{tab:mit}).
Our approach, without using pretraining, outperforms all state-of-the-art methods. 
We further note that AIRStreams outperforms the powerful AssembleNet~\cite{ryoo2020assemblenet} which additionally pre-trained from the large Kinetics video dataset, whereas our semantics stream is obtained from a non-related image dataset. For direct comparison we also show (RGB+Flow)-only AIRStreams, which outperforms 4-stream AssembleNet~\cite{ryoo2020assemblenet} when both are trained from scratch (33.53 for AIRStreams vs 31.41 for AssembleNet); (RGB+Flow)-only AIRStreams without pre-training also outperforms the other 2-stream methods. 

\vspace{-0.3cm}
\subsubsection{Charades dataset}
 Charades dataset is also a challenging dataset of longer video sequences in which multiple labels are provided per video clip, i.e., a multi-class multi-label setting. We use the standard evaluation protocols, and since the data is small compared to the number of classes available, we use MiT pre-training.
AIRStreams works very well (Table~\ref{tab:charades}) compared to the prior state-of-the-art on this dataset which also used pre-training. It outperforms all methods, including SlowFast~\cite{feichtenhofer2018slowfast} and video architecture search EvaNet~\cite{piergiovanni2018evolving}, with the 
exception of the powerful 4-stream AssembleNet with connection learning between streams~\cite{ryoo2020assemblenet}, which on this dataset performs best. 
This may be due to the longer duration of videos in Charades for which the architecture searched connectivity of AssembleNet is more beneficial. 

\begin{table} [t]
\begin{center}
\adjustbox{width=0.99\linewidth}{
\begin{tabular}{lc}
\toprule
Method &   mAP(\%) \\                         
\midrule
\makecell[l]{BatchNorm Inception~\cite{ioffe2015batchnorm}}  & 11.6 \\ 
TSN-Flow~\cite{wang2016temporal}  & 15.7  \\  
\makecell[l]{Two-stream \cite{simonyan2014two} (from \cite{sigurdsson2016asynchronous})}   & 18.6 \\
   Asyn-TF \cite{sigurdsson2016asynchronous} & 22.4 \\
    CoViAR \cite{wu2018compressed}  & 21.9 \\
    TRN~\cite{zhou2018temporal}   & 25.2 \\
    I3D \cite{carreira2017quo}  & 32.9 \\
    I3D \cite{carreira2017quo} (from \cite{wang2018non})  & 35.5 \\
 
   EvaNet \cite{piergiovanni2018evolving}    & 38.1 \\
   SlowFast \cite{feichtenhofer2018slowfast}  & 45.2 \\
    Two-stream (2+1)D ResNet50 (Baseline)  & 46.5 \\
   AssembleNet~\cite{ryoo2020assemblenet} (MiT pretraining)  & \textbf{53.0} \\
\midrule
AIRStreams (Ours, MiT pretraining)    & 50.1                      \\  
\bottomrule
\end{tabular}}
\caption{Performance on Charades. AIRStreams is able to outperform other methods with the exception of powerful AssembleNet, which uses architecture search for learning connectivity.}
\label{tab:charades}
\end{center}
\vspace{-4mm}
\end{table}

\begin{table} []
\begin{center}
\adjustbox{width=0.99\linewidth}{
\begin{tabular}{lcccc}
\toprule
Method & \makecell[c]{Uses\\pose} & \makecell[c]{Uses\\pretr.} & \makecell[c]{Classif.\\\%}& \makecell[c]{~Mean\\per-cls. \%~} \\
\midrule
LSTM \cite{mahasseni2016lstm} & Yes & No & - & 42.5 \\
\makecell[l]{Dense Trajectories \cite{deng2011dt}} & No & No & - & 41.9 \\
\makecell[l]{I3D \cite{carreira2017quo}} & No & Yes & 72.0 & 53.4 \\
\makecell[l]{NonLocal \cite{wang2018non}} & No & Yes & - & 53.6 \\
\makecell[l]{Separable STA \cite{das2019toyota}} & Yes & Yes & 75.3 & 54.2 \\
\midrule
\makecell[l]{AIRStreams (Ours)} & No & No & \textbf{78.28} & \textbf{62.11}  \\
\bottomrule
\end{tabular}}
\caption{Classification and mean per-class accuracies on Toyota Smarthome. The pose column indicates whether the precomputed pose was used and the pretraining column indicates whether pretraining on another video dataset was used. AIRStreams \textbf{does not use pre-training}. Some prior methods are reported by~\cite{das2019toyota}.}
\label{tab:toyota}
\vspace{-8mm}
\end{center}
\end{table}

\vspace{-3mm}
\subsubsection{Toyota Smarthome dataset}
\label{sec:toyota}
We further evaluate on the Toyota Smarthome dataset~\cite{das2019toyota} which contains activities within peoples' homes. We use the standard protocol in prior work~\cite{das2019toyota} and report both the activity classification accuracy (\%) and the `mean per-class' accuracy (\%). Table~\ref{tab:toyota} shows the results for the more challenging `Cross-Subject' (CS) evaluation setting.
As seen, AIRStreams, without pretraining, outperforms notably previous work which used pre-training (some from Kinetics) and additional 3D human joints information~\cite{das2019toyota}. More specifically, it outperforms the best prior work by 3\% and 8\% for the classification and mean per-class accuracy, respectively, which indicates it is performing very well on under-represented classes.

\section{Conclusion}
We present the `AIRStreams' approach for video understanding which introduces semantic segmentation as an intermediate representation, alongside with flow, and jointly optimizes both the representations themselves and their importance for the final task. 
AIRStreams extends the current 2-stream models. By incorporating semantics and training jointly, it outperforms or is competitive to the best SOTA models SlowFast and AssembleNet.
Our work may encourage researchers to incorporate 
other intermediate representations, which will help tackle even more challenging and more fine-grained 
activity recognition tasks.



\balance
{\small
\bibliographystyle{ieee_fullname}
\bibliography{bib}
}

\newpage
\appendix
\nobalance
\section{Additional visualizations}

\begin{figure*}[t]
    \centering
    \includegraphics[width=\textwidth]{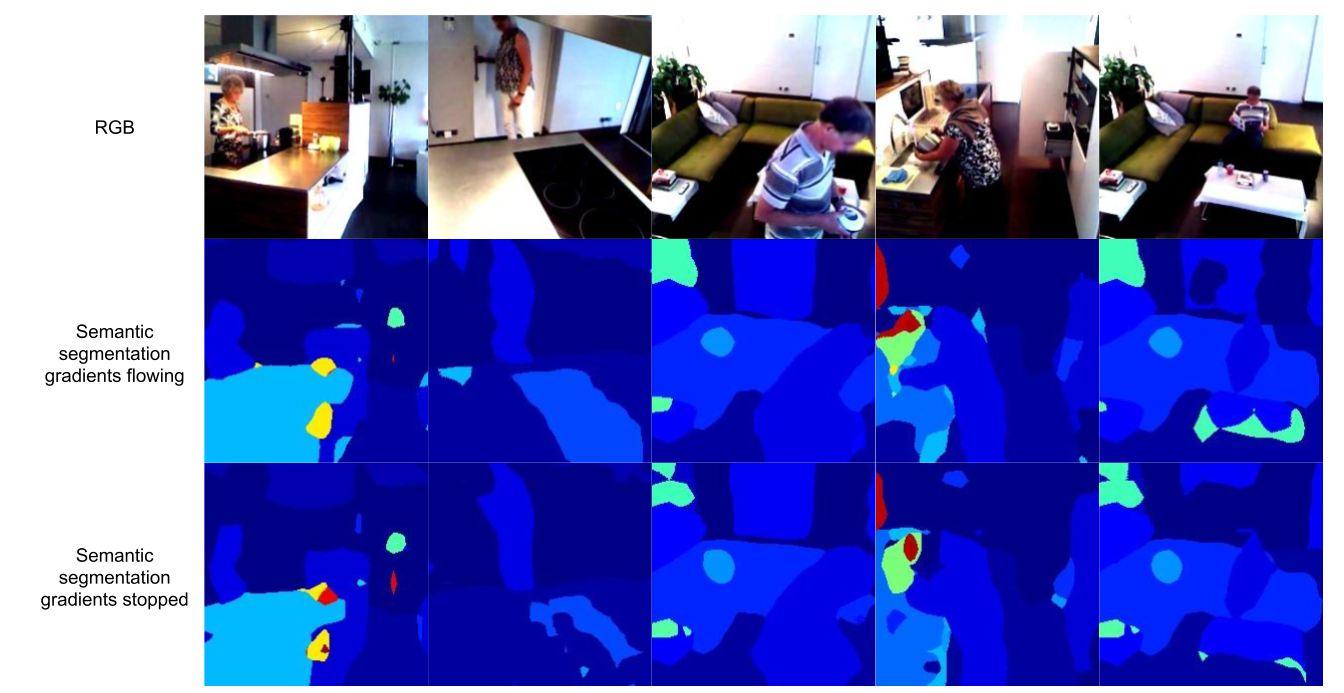}
    \caption{Predictions of the semantic segmentation mini-network. First row shows an RGB input frame. Second row shows a semantic segmentation prediction from a network where gradients were flowing from the video classification losses all the way through the semantic segmentation task. Third row shows a semantic segmentation prediction from a network where video classification gradients were stopped at the semantic segmentation logits.}
    \label{fig:semantics_visualization}
\end{figure*}

In this section we show visualizations of the intermediate representations. Figure~\ref{fig:semantics_visualization} shows the semantic segmentation intermediate representation both with and without gradients flowing for the Toyota Smarthome dataset.

\section{Network architecture}

\begin{figure*}
    \centering
    \includegraphics[width=\textwidth]{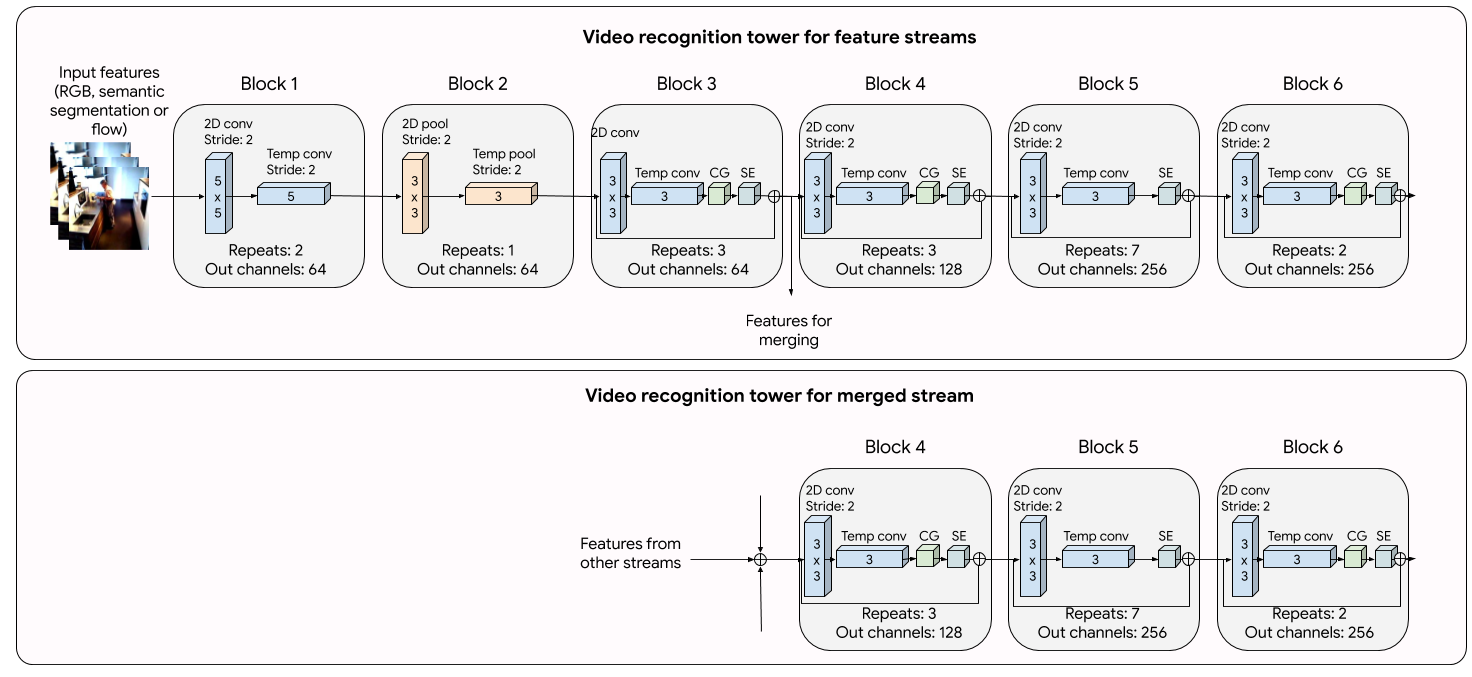}
    \caption{AIRStreams video recognition tower architecture. Each individual feature tower (RGB, flow and semantic segmentation) is an instance of the network displayed in the top section. The tower for the merged stream is displayed in the bottom section and takes as input the features from Block 3 of all the feature streams. The merged stream tower is identical to the Blocks 4-6 of the feature streams. The network is built from six blocks, each repeated one or more times. Each block is assembled from the following layers: 2D convolution over the spatial dimensions (2D conv), 1D convolution over the temporal dimension (Temp conv), max-pooling over either the spatial or temporal dimensions (2D pool, Temp pool), context-gating layer (CG), and squeeze-excite (SE) layer. The kernel size of each layer is written inside the illustrating box. Skip connections over each block are illustrated with connecting lines.}
    \label{fig:estreams_tvn}
\end{figure*}

In Figure~\ref{fig:estreams_tvn} we visualize the network architecture which is simple, yet efficient.

\section{Evolutionary search progression}

\begin{figure*}[t]
    \centering
    \includegraphics[width=0.7\textwidth]{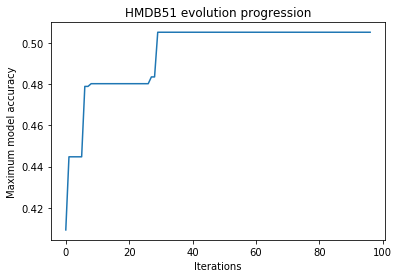}
    \caption{Accuracy of the best found model as a function of the number of iterations.}
    \label{fig:evolution_progression}
\end{figure*}

In Figure~\ref{fig:evolution_progression} we visualize the progression of the evolutionary search for the optimal loss weights. The graph shows the accuracy of the best found model as a function of total iterations, i.e. models evaluated. The evolution was done on the HMDB-51 dataset (without pretraining). As seen, not too many iterations are needed to converge to a good solution.

\end{document}